\documentclass{article}
\usepackage{graphicx} 
\usepackage[margin=1.0in]{geometry} 
\usepackage{float}
\usepackage{csvsimple}

\title{A Dataset for Mechanical Mechanisms}

\author{
    Farshid Ghezelbash\thanks{Department of Computer Science, Georgia Institute of Technology, Atlanta, GA, USA. Corresponding author. Email: \href{mailto:ghezelbash.far@gmail.com}{ghezelbash.far@gmail.com}} \and
    Amir H Eskandari\thanks{Division of Applied Mechanics, Department of Mechanical Engineering, Polytechnique Montréal, Canada.} \and
    Amir J Bidhendi\thanks{Department of Computer Science, Georgia Institute of Technology, Atlanta, GA, USA.}
}

\date{August 2024}

\begin{document}

\maketitle

\begin{abstract}
\noindent This study introduces a dataset consisting of approximately 9,000 images of mechanical mechanisms and their corresponding descriptions, aimed at supporting research in mechanism design. The dataset consists of a diverse collection of 2D and 3D sketches, meticulously curated to ensure relevance and quality. We demonstrate the application of this dataset by fine-tuning two models: 1) Stable Diffusion (for generating new mechanical designs), and 2) BLIP-2 (for captioning these designs). While the results from Stable Diffusion show promise, particularly in generating coherent 3D sketches, the model struggles with 2D sketches and occasionally produces nonsensical outputs. These limitations underscore the need for further development, particularly in expanding the dataset and refining model architectures. Nonetheless, this work serves as a step towards leveraging generative AI in mechanical design, highlighting both the potential and current limitations of these approaches.

\end{abstract}

\section{Introduction}
Mechanical mechanism design has traditionally followed a structured approach, beginning with engineers identifying a problem, followed by exploring and analyzing existing mechanisms to directly apply to or inspire solutions. This method, while effective, can be time-consuming and limited by the scope of known mechanisms. We propose that generative AI models hold remarkable potential for a paradigm shift in how mechanical mechanisms are conceived. These AI tools can assist in generating novel mechanism ideas or aid in the brainstorming phase, potentially expediting and streamlining the design process. Although some text-to-image generation tools, such as DALL-E and Midjourney, are already available, these tools are generally not fine-tuned for mechanical systems. Consequently, they often generate outputs with limited relevance for engineering design, underscoring the need for specialized datasets that can be used for fine-tuning or training these models. 

This work presents a dataset of ~9,000 images of mechanical mechanisms (2D and 3D sketches), each accompanied by a text description. To evaluate the utility of this dataset, we used it in training models which were then used in generating new mechanism and in developing image captioning models. This work provides a specialized dataset for the research community, and illustrates the potential of AI-driven approaches in advancing mechanical design processes.
\section{Methods}
The dataset of mechanical mechanisms was compiled through web scraping from various sources. The first source was a YouTube channel focused on mechanical design (Table \ref{table:datasets}), where mechanisms were primarily modeled using CAD software \cite{thang010146_channel}. For each mechanism, we extracted a frame from the video along with its description. The second source was a digital library dedicated to mechanisms and gears \cite{dmg_lib_video}, which included a  video section featuring 3D reconstructions. These videos provided detailed visualizations of mechanisms, from which we extracted images and descriptions (Table \ref{table:datasets}). The third source was a book that contains a vast collection of 2D sketches and comprehensive descriptions of mechanical mechanisms \cite{artobolevsky1975mechanisms}. In total, these sources yielded 8,994 images and corresponding descriptions.

To ensure the quality of the dataset, we conducted a thorough manual review of all the images. During this process, we identified and removed any images that were blank, irrelevant, or did not make sense. To clean up descriptions, we manually edited them to ensure consistency and relevance. We used ChatGPT to remove references to patents/designs, creators' names, or other verbose and/or unrelated content by providing specific instructions to remove any such references, including mentions of detailed variants or referrals to other designs. 
\begin{table}[h!]
\centering
\caption{Summary of datasets with the number of images, and references.}
\begin{tabular}{|c|c|c|}
\hline
 Number of Images & Sketch Type & Reference \\ \hline
 3872               & 3D          & \cite{thang010146_channel}\\ \hline
 980& 3D          & \cite{dmg_lib_video} \\ \hline
 4142             & 2D          & \cite{artobolevsky1975mechanisms} \\ \hline
              Total: 8994     &             &           \\ \hline
\end{tabular}
\label{table:datasets}
\end{table}
\subsection{Textual Data Analysis and Visualization}
To analyze the text descriptions of mechanisms, we processed the dataset by tokenizing the text and removing common stop words \cite{nlp}. The frequency of the remaining words was calculated using a word counter to identify key terms. A word cloud was generated to visually represent the most frequent words, and the top words were further analyzed to identify potential synonyms using WordNet, a lexical database that groups English words into sets of synonyms.

\subsection{Design Generation Using Stable Diffusion}
To evaluate the application of our dataset, we used it to fine-tune Stable Diffusion 1.6 \cite{stableDiff} which is a deep learning model known for its ability to generate high-quality images from textual descriptions. For our purpose, we utilized the entire dataset of 8,994 images and descriptions to fine-tune this model specifically for generating mechanical mechanism designs. The fine-tuning process allowed the model to learn the characteristics and visual features of the mechanical systems.

\subsection{Captioning Using BLIP-2}
To further our evaluation of the application of this dataset, we fine-tuned the BLIP-2 (Bootstrapped Language-Image Pretraining) model using our database \cite{blip}. BLIP-2 is a vision-language model designed to generate descriptive captions for images by understanding the relationship between visual content and textual information. Fine-tuning BLIP-2 on our dataset aimed to improve its ability to produce accurate and concise descriptions of mechanical mechanisms and to ensure that the captions generated were relevant, technically appropriate, and conducive to the usability of the dataset for further research in mechanical design.

\section{Results and Discussion}
\begin{figure}[H]
    \centering
    \includegraphics[width=0.6\linewidth]{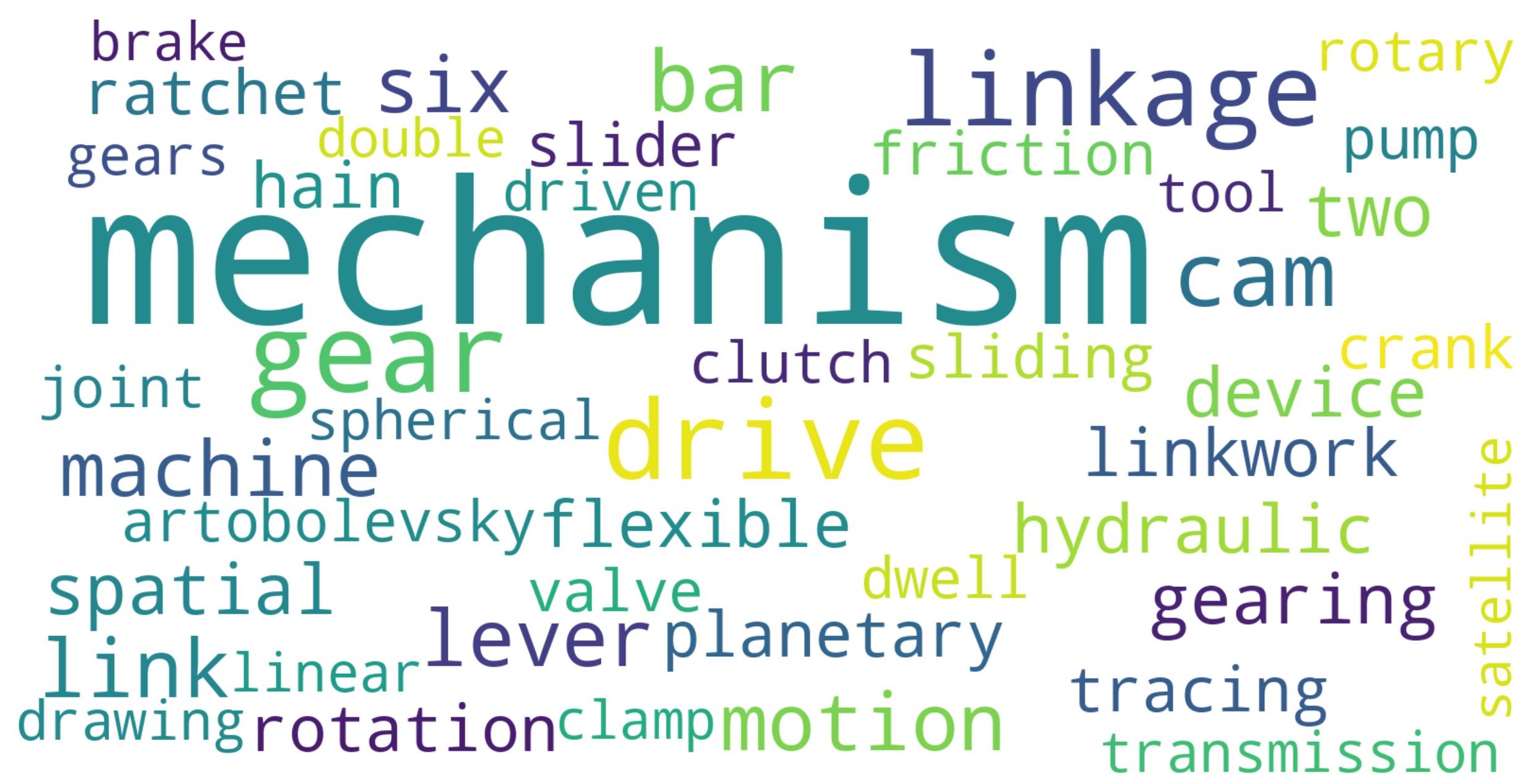}
    \caption{Word cloud of text description of mechanisms.}
    \label{fig:world_cloud}
\end{figure}

\begin{figure}[H]
    \centering
    \includegraphics[width=0.85\linewidth]{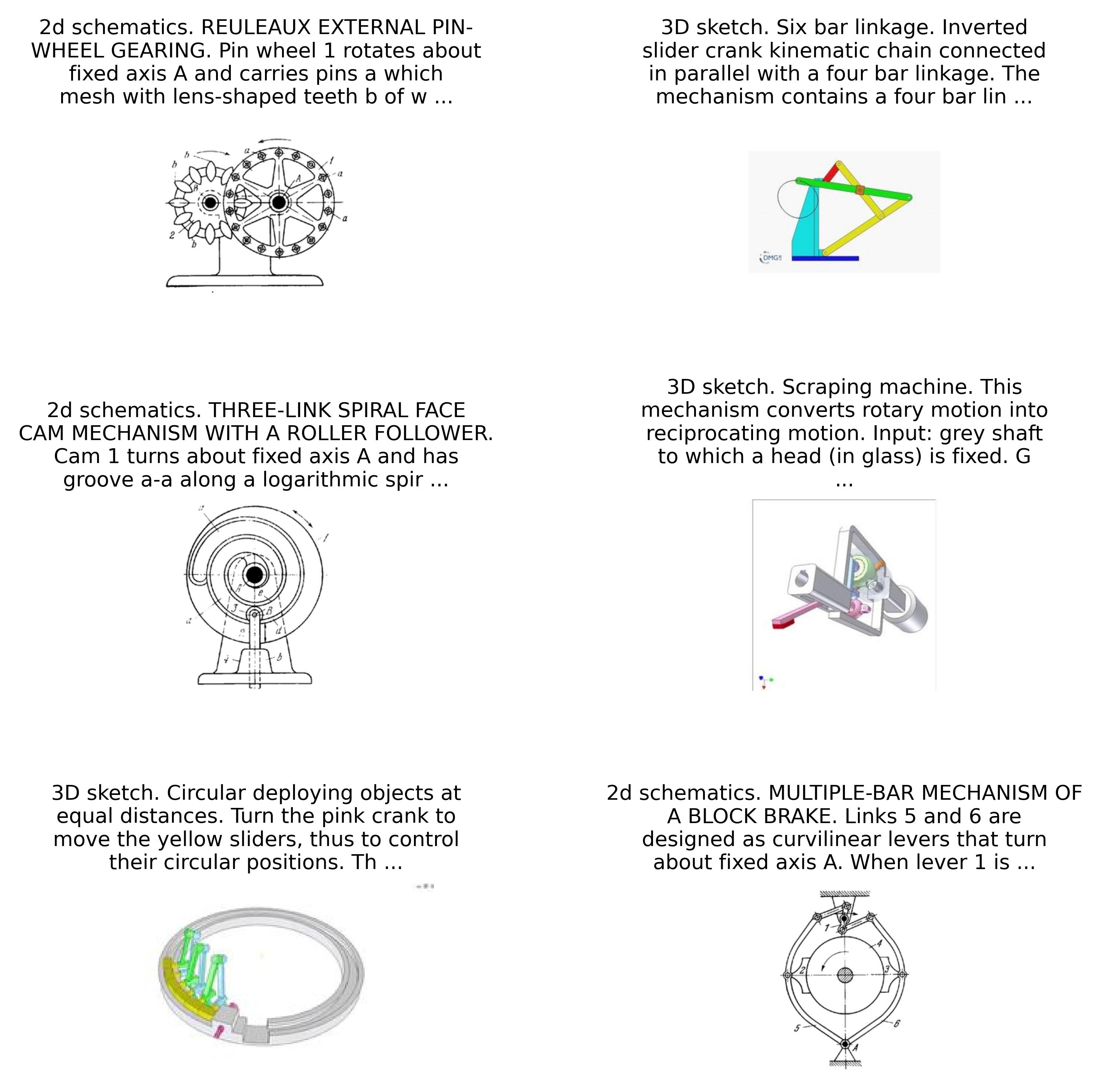}
    \caption{Nine randomly selected mechanisms with their descriptions (limited to 150 characters).}
    \label{fig:sample_data}
\end{figure}

After processing and cleaning the datasets (Table \ref{table:datasets}), we composed a collection of 8,994 images with their corresponding descriptions. This dataset, while relatively modest in size, provides a foundational resource for training/fine-tuning models in the context of mechanical mechanism design. The word cloud in Figure \ref{fig:world_cloud} visually represents the frequency of terms used in the descriptions, highlighting key concepts and components that are prevalent in mechanical mechanism design. Figure \ref{fig:sample_data} showcases a sample of nine randomly selected mechanisms along with their descriptions, which illustrate the diversity of the mechanical mechanisms included in the dataset. Additionally, Table \ref{table:top_words} summarizes the most frequent key terms and their synonyms, offering an insight into the linguistic patterns in the dataset.

\begin{table}[H]
\centering
\caption{Frequency of key terms and their synonyms}
\label{table:top_words}
\begin{filecontents*}{top_words.csv}
Index,Word,Frequency,Synonyms
1,mechanism,4850,mechanism; chemical mechanism; mechanics
2,gear,541,paraphernalia; gearing; gear mechanism; appurtenance; ...
3,drive,533,motor; repulse; beat back; parkway; driving; ride; force; repel; ...
4,linkage,451,gene linkage; linkage
5,cam,362,Cam River; cam; River Cam; Cam
6,bar,318,saloon; streak; barricade; Browning automatic rifle; debar; bar; ...
7,link,283,radio link; contact; colligate; tie in; tie; unite; connect; relate; ...
8,motion,279,motion; apparent motion; motility; gesture; apparent movement; ...
9,lever,272,jimmy; prise; pry; prize; lever; lever tumbler
10,six,268,half dozen; half dozen; Captain Hicks; 6; vi; six spot; half a dozen; ...
\end{filecontents*}
\csvautotabular{top_words.csv}
\end{table}

\subsection{Design Generation Using Stable Diffusion}
\begin{figure}[H]
    \centering
    \includegraphics[width=0.75\linewidth]{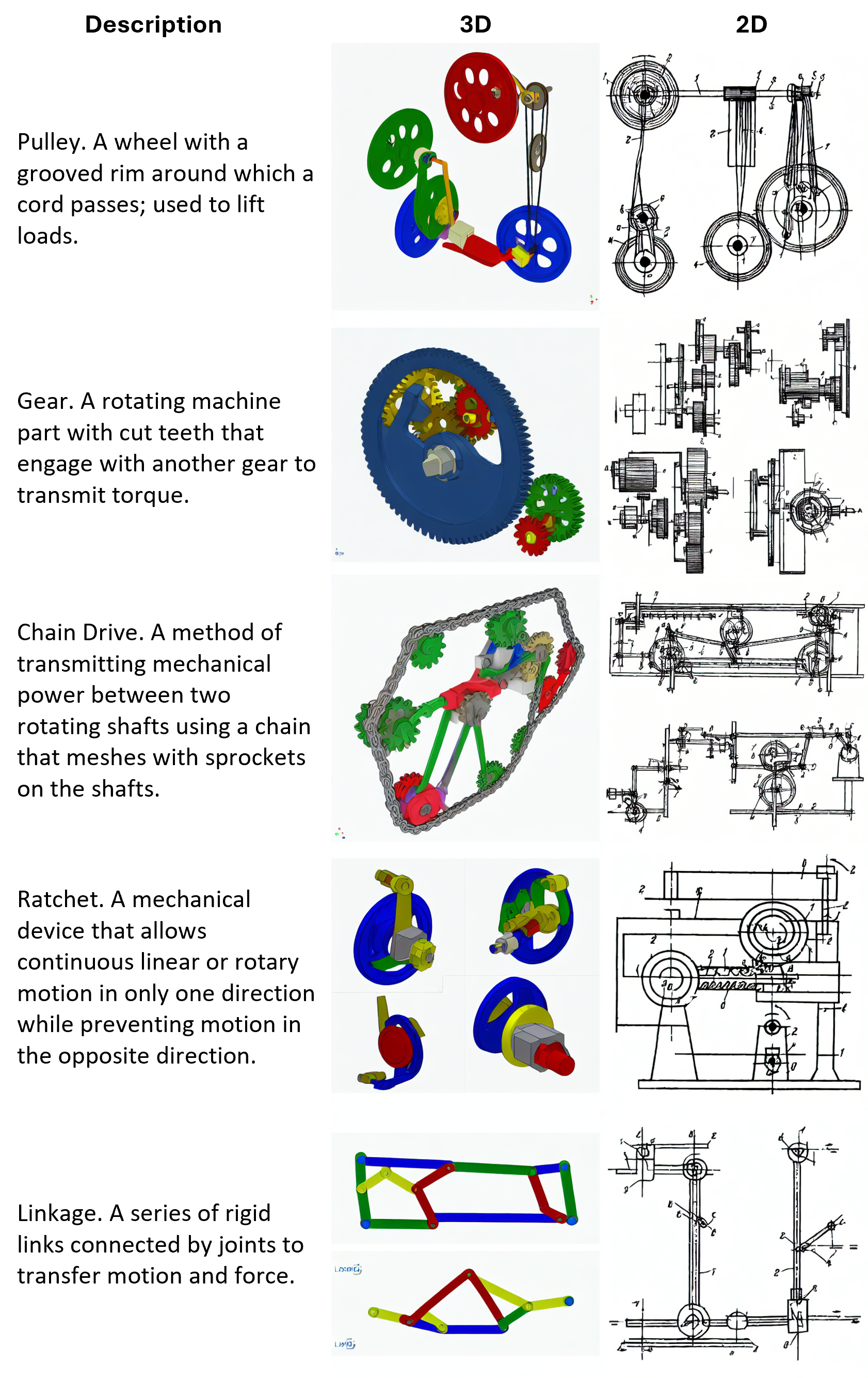}
    \caption{Examples of 2D (right) and 3D (middle) sketches generated by the fine-tuned Stable Diffusion model from a text input (left). The 3D sketches generally align better with the provided descriptions, capturing key components of mechanical mechanisms, while the 2D sketches often lack coherence and meaningful structure.}
    \label{fig:gen_mech_notBad}
\end{figure}

\begin{figure}[H]
    \centering
    \includegraphics[width=0.75\linewidth]{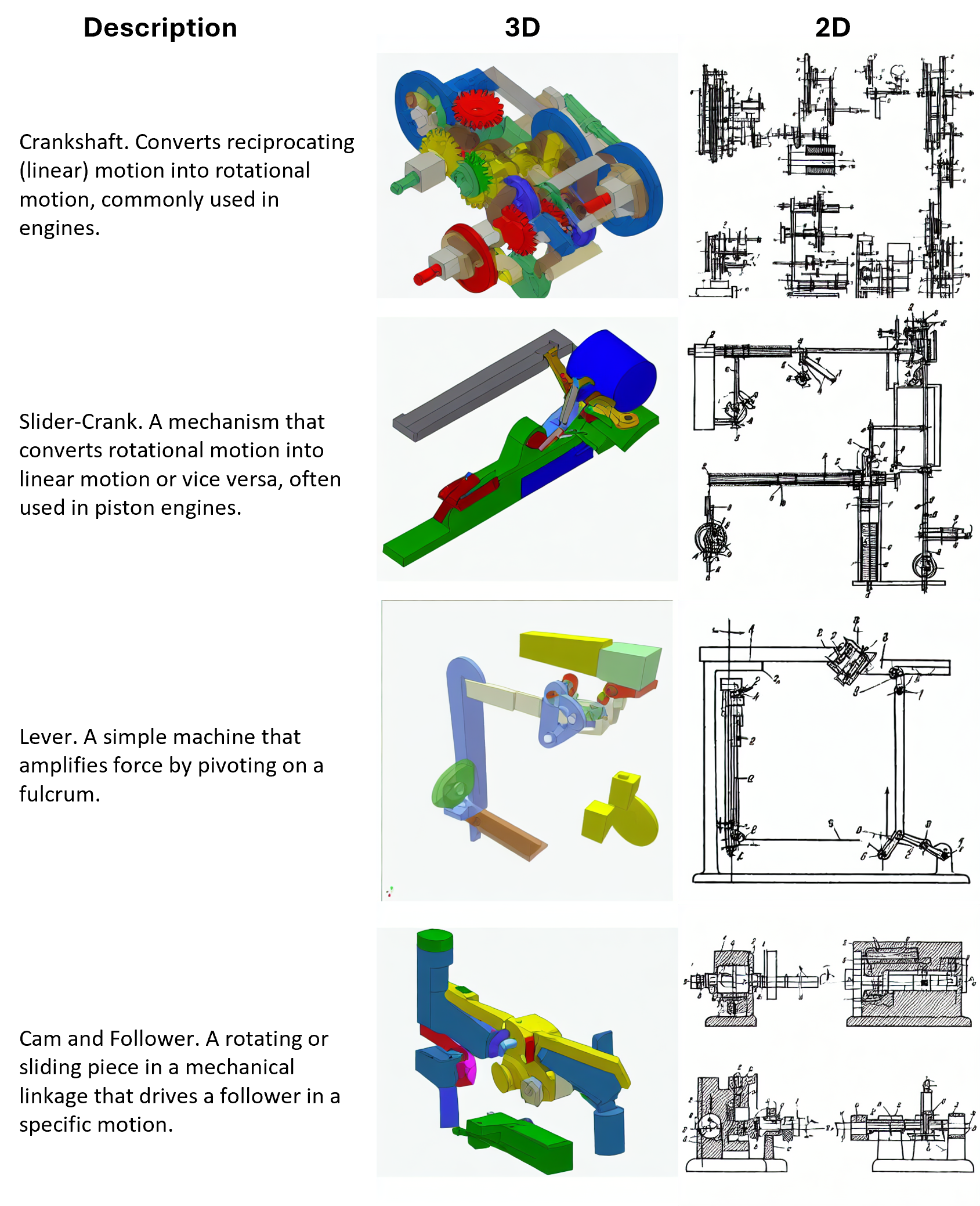}
    \caption{Examples of nonsensical/hallucinated outputs generated by the fine-tuned Stable Diffusion model from a text input (2D: right; 3D: middle; input prompt: left). These examples illustrate that the model occasionally struggles to accurately interpret prompts, resulting in outputs that do not align with the intended mechanical designs and lack meaningful structure.}
    \label{fig:gen_mech_hall}
\end{figure}

After fine-tuning the model (1500 epochs) for Stable Diffusion model on the presented dataset, we generated both 2D and 3D sketches (by adding "2D schematic" or "3D sketch" at the beginning of each prompt; Figure \ref{fig:gen_mech_notBad} and Figure \ref{fig:gen_mech_hall}). We noted that some of the generated mechanisms made relatively good sense (Figure \ref{fig:gen_mech_notBad}), particularly the 3D sketches, which often included essential components of the provided descriptions. However, the 2D sketches in most examples were relatively meaningless and lacked coherence (Figure \ref{fig:gen_mech_notBad} and Figure \ref{fig:gen_mech_hall}), which indicates the great potential of this database and models, but also highlights significant room for improvement. During experimentation, we also noticed that the model occasionally produced nonsensical outputs, especially for certain prompts where it tended to hallucinate and generate figures that were not only inaccurate but also entirely unrelated to the intended design (Figure \ref{fig:gen_mech_hall}). This tendency indicates the need for further refinement of the model, particularly in handling more complex or specific prompts to reduce instances of such errors and improve the overall reliability of the generated sketches.

\subsection{Captioning Using BLIP-2}
\begin{figure}[H]
    \centering
    \includegraphics[width=1.0\linewidth]{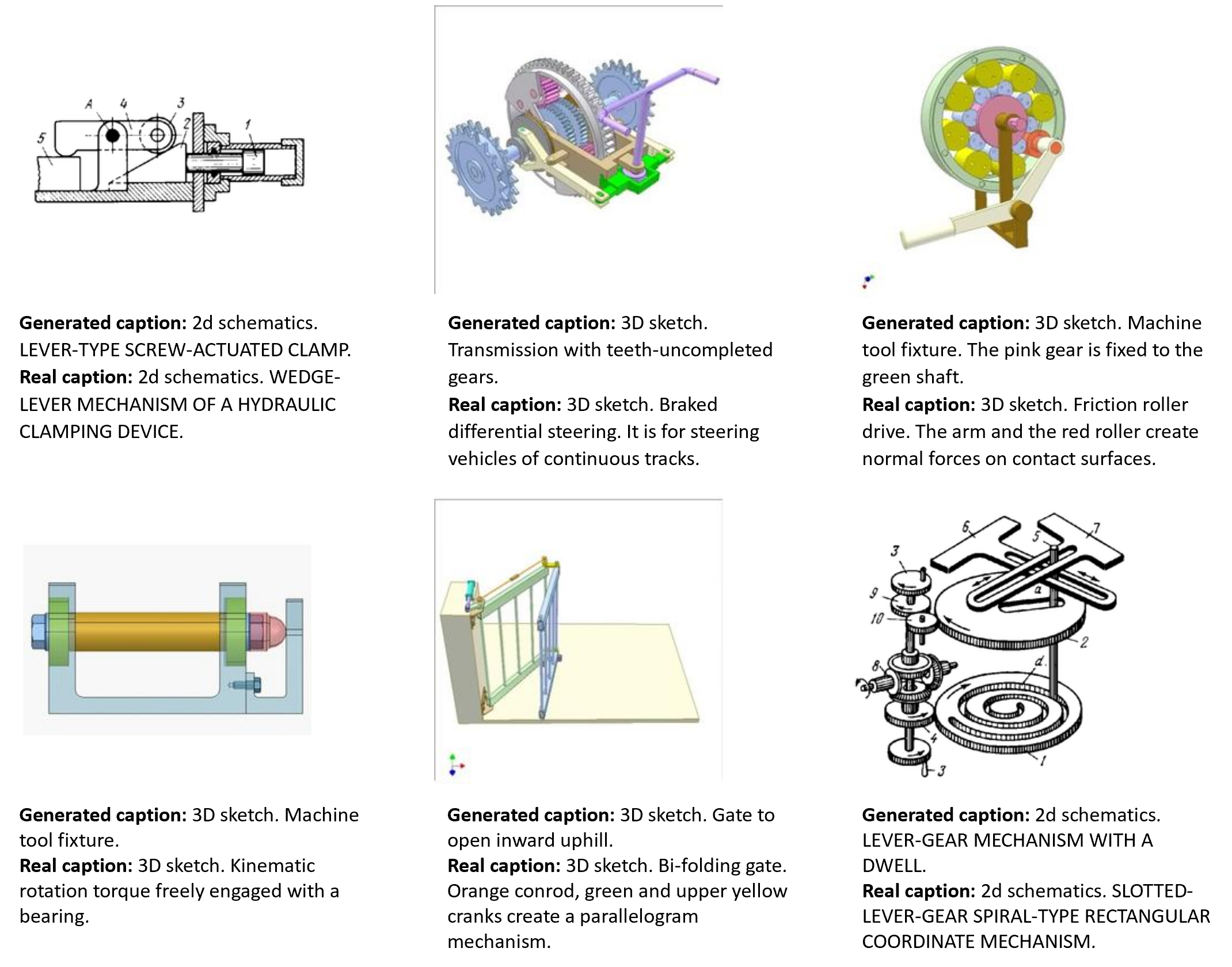}
    \caption{Six randomly selected mechanisms with their model-generated and real captions.}
    \label{fig:caption}
\end{figure}

The results of the BLIP-2 model fine-tuning for captioning mechanical mechanisms are mixed. Most of the generated captions are incorrect, with only a few containing elements of truth. This inconsistency is likely due to the limited number of training epochs (=10), which is far from sufficient for achieving accurate results. Our primary goal was to demonstrate the potential of this approach, despite significant limitations in training resources, particularly GPU access. The generated captions, along with their corresponding images, are presented in Figure \ref{fig:caption}.

\subsection {Limitations}
One of the primary limitations of this study is the relatively small size of the dataset, which includes only 8,994 images and descriptions. However, we believe this is a foundational step toward the use of generative models in mechanism design. This modest size of the dataset limited the model's ability to generalize across a broader range of mechanical designs, resulting in less reliable outputs, particularly for more complex or novel inputs.

\subsection {Future Directions}
\begin{itemize}
    \item Expand the dataset by incorporating a wider variety of mechanical mechanisms, including more complex and diverse designs.
    \item Refine the model's architecture and training procedures to reduce the occurrence of nonsensical outputs and improve the coherence of 2D sketches.
    \item Explore alternative generative models or integrate multiple models to enhance the quality of the generated designs.
    \item Apply the model to real-world design challenges and iteratively improve its performance based on feedback from engineering professionals, transitioning the research from theoretical to practical application.
\end{itemize}

\section{Code and Dataset Availability}
\sloppy
The code used for fine-tuning the models and generating the results presented in this paper is available on GitHub: \url{https://github.com/farghea/database_for_mechanical_mechanism}. The dataset, including 256x256 versions (\url{https://drive.google.com/file/d/1yC6nKih8HcAAoKCVM-Lo6bxGQ2O8T5-_/view?usp=sharing}) and higher resolution (\url{https://drive.google.com/file/d/1jqSKDypbN3vfGBA2SnUuQLuSnZC3BPYh/view?usp=sharing}), can be accessed from Google Drive.
\sloppy

\section{Responsibility of the Use of the Dataset}
All data included in this dataset was collected from publicly available sources on the internet. We have ensured that the data was freely accessible at the time of collection. To respect and acknowledge the contributions of the original creators, users of this dataset are strongly encouraged to cite the corresponding references provided in the dataset documentation. It is the responsibility of the users to ensure that the dataset is used ethically and in accordance with any applicable laws or regulations.

\bibliographystyle{plain}
\bibliography{bibFile}

\end{document}